# Omni-directional Feature Learning for Person Re-identification

Di Wu[1], Hong-Wei Yang[1] and De-Shuang Huang[1*]

[1]Institute of Machine Learning and Systems Biology, School of Electronics and Information Engineering, Tongji University, Caoan Road 4800, Shanghai 201804, China

*Abstract—* Person re-identification (PReID) has received increasing attention due to it is an important part in intelligent surveillance. Recently, many state-of-the-art methods on PReID are part-based deep models. Most of them focus on learning the part feature representation of person body in horizontal direction. However, the feature representation of body in vertical direction is usually ignored. Besides, the spatial information between these part features and the different feature channels is not considered. In this study, we introduce a multi-branches deep model for PReID. Specifically, the model consists of five branches. Among the five branches, two of them learn the local feature with spatial information from horizontal or vertical orientations, respectively. The other one aims to learn spatial information between the different feature channels generated by the last convolution layer. The remains of two other branches are identification and triplet sub-networks, in which the discriminative global feature and a corresponding measurement can be learned simultaneously. All the five branches can improve the representation learning. We conduct extensive comparative experiments on three PReID benchmarks including CUHK03, Market-1501 and DukeMTMC-reID. The proposed deep framework outperforms many state-of-the-art in most cases.

*Keywords— person re-identification, deep learning, triplet model, identification model, GRU.*

## I. INTRODUCTION

As a fundamental task of intelligent surveillance, person re-identification (PReID) is to re-identify a specific pedestrian from multiple non-overlapping camera views. It has been of considerable interest to the computer vision community in recent years. Great progress has been made in PReID, however, it remains a challenge issue for the visual appearance may undergo significant variations when facing unpredictable changes in illumination, background clutter as well as person pose.

In current researches, PReID is resolved from the following two angles: 1). Extracting discriminative features to represent different identities. 2). Learning an effective distance metric to make the relative distance between the inter-class larger than intra-class.

Benefiting from the great development of deep learning technology in computer vision, a large number of deep architecture-based methods have been introduced for PReID. Different from traditional hand-craft methods, these deep learning-based methods integrate the feature and distance metric learning by an end-to-end way. It is worth mentioning that the most recent state-of-the-art results are all achieved by deep learning-based models. Many deep learning-based models learn global pedestrian feature. When the global features of pedestrians are obtained, the Euclidean metric is applied to measure the distance between two pedestrians. Global feature learning methods have the following drawbacks: a) the irrelevant information may be introduced in global feature when pedestrian's body facing occlusion, b) global features are not enough discriminating to represent the pedestrians that with similar appearances. To alleviate these dilemmas above, some studies have tried to locate body parts to learn local features for pedestrians. Some works use predefined horizontal stripes to divide the body into several partitions, and separately use the partitions to learn part features. However, it suffers from alignment problem under the occlusion and imperfect detection condition. Other works introduce pose annotation information to address the alignment issue, which requires extra pose estimation procedure. Moreover, most of these part-based methods do not take into account the spatial contextual information between the different divided parts. In [1], the authors first use Recurrent Neural Network (RNN) to exploit the spatial context information between the extracted sequence features. However, the processes of spatial information learning and feature extraction in this work are separate. Bai et al. [2] propose to apply Long Short-Term Memory (LSTM) to learn the spatial contextual information between different body parts from head to foot. Yet, the spatial contextual relationship between body parts from horizontal orientation, i.e. left to right, is ignored. It is noteworthy that almost all part-based models ignore to learn part features of body from left to right. However, the part features from left to right may be very useful, especially when the left or right side of body is occluded. As shown in Figure 1, the left or right side of bodies are occluded, and using the part features from head to foot can introduce irrelevant information. Then the part features from left to right are useful in this situation. Therefore, we propose to adopt gated recurrent unit (GRU) to simultaneously learn the spatial information between different body parts from head to foot as well as from left to right.

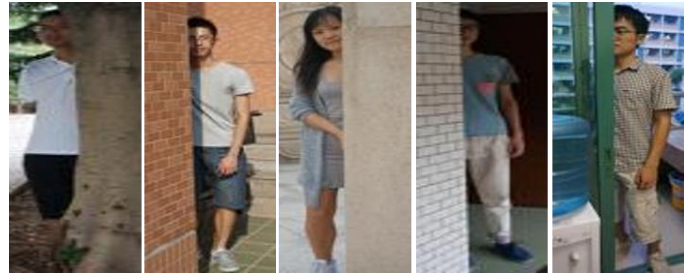

**Figure 1.** Left or right side of bodies is occluded. Using part features from head to foot may introduce irrelevant information.

* Corresponding author. E-mail address: dshuang@tongji.edu.cn

Recent works have shown that the robust property of convolutional neural network (CNN) features can be enhanced by integrating some additional learning module that help capture the correlations between channels. In this study, apart from part-based feature, we also investigate the interdependencies between the channels of its deep feature maps to improve the discriminatively ability of the learned representations. For this purpose, we use the generated feature maps to input GRU model to learn the interdependencies knowledge of them.

Some works have shown that the combination of triplet loss and identification loss can promote the quality of the learned deep feature. Triplet loss makes the Euclidean distances between the negative pairs larger than that of positive pairs. The training target of triplet loss is similar to its test manner, however, it uses weak label information. Identification loss regards the PReID tasks as a multi-class classification issue, thus, it makes use of fully annotation information. Yet, the training goal of identification loss is different from its test manner. Therefore, the two types of loss functions have complementary advantages and limitations. In this study, we also adopt this hybrid strategy. Unlike previous works, we utilize Online Instance Matching (OIM) loss [3] rather than commonly used Softmax loss for identification task. The generalization ability of Softmax loss from training set to test set may weak when dataset contains a large number of identities and each identity only has limited number instances. One possible reason for this is that the Softmax loss has to learn too much discriminant functions with limited instances for each identity. Thus, the classifier matrix cannot be fully learned at each back propagation stage. Compared to it, OIM loss is nonparametric, therefore, the gradients are performed on the features rather than classifier matrix directly. Our previous work [4] also proves that the effectiveness of the combination of OIM and triplet losses. As shown in Figure 2, we use the features generated by fully connection layer and part-based as well as channels features produced by GRU model to input the identification subnetworks, respectively. Moreover, the triplet loss is applied on the pooled features for learning a corresponding similarity measurement. In the test phase, we choose features $f_{OIM}$ as the final pedestrian descriptors for Market-1501 and DukeMTMC-reID datasets. As for the CUHK03, we use features $f_{Trip}$ as the final descriptors.

In summary, the contributions of this study are:

1) We propose to use GRU to learn the omni-directional correlation information for pedestrian body. The proposed model not only consider the spatial information between different body parts from head to foot, but also from left to right.

2) We propose to learn the interdependencies knowledge between channels to enhance the discriminatively ability of the learned descriptors.

3) Our proposed method outperforms other competitive state-of-the-art methods on the widely used PReID datasets.

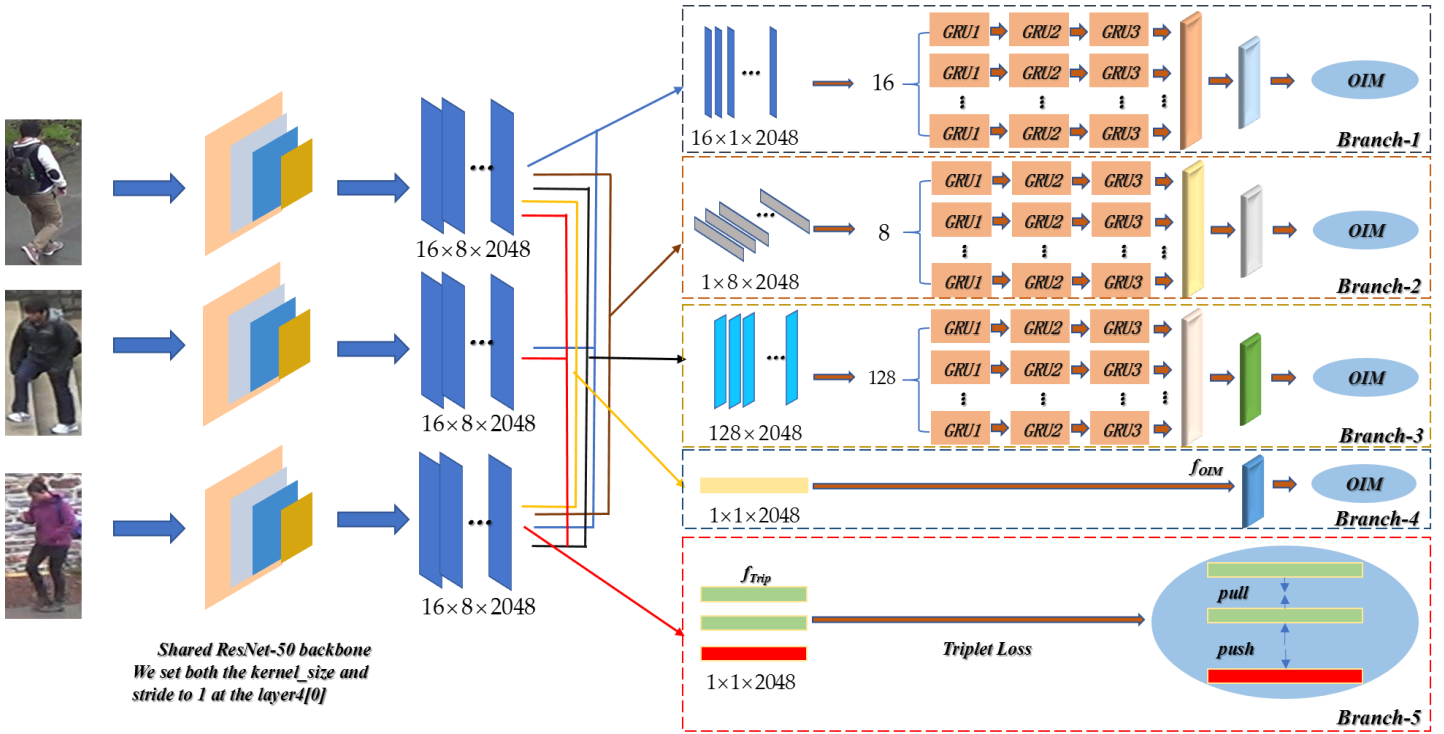

**Figure 2.** Illustration of the proposed deep model. We use ResNet-50 [5] as backbone network and set **both the kernel size and stride to 1 at the layer4 [0]** of the pre-trained ResNet-50 model. The architecture contains five branches. The above two branches learn the part-based features of body from horizontal and vertical orientations, respectively. The middle branch learns the interdependencies knowledge between channels. The bottom two branches learn the global features and a corresponding measurement for PReID. The whole deep architecture is supervised by four OIM loss functions and one triplet loss function. The picture best viewed in color.

## II. RELATED WORK

With the development of deep learning technology, especially convolutional neural network (CNN), deep feature learning by CNN has become a frequent method in PReID domain. Most of deep learning-based structures are based on the following models: identification model, verification model and triplet model. Identification-based model treats the PReID as a task of multi-classification issue. Verification-based model takes a pair of images as input and outputs a similarity value to determine whether the paired images are the same pedestrian or not. Triplet-based model aims to make the distances between the same person images as small as possible while make the distances between different person images as large as possible. Some approaches combine two types of models mentioned above. Chen et al. [6] design a multi-task deep network that integrates the triplet and verification losses to take the advantage of the two losses for person ReID. Wang et al. [7] analyzes the advantages and limitations of single-image representation (SIR) and the classification of cross-image representation (CIR) in PReID community. They propose pairwise comparison and triplet comparison formulations to simultaneously learn the CIR and SIR. Qian et al. [8] introduce a multi-scale deep architecture which contains verification and classification subnets for PreID. In this study, we employ OIM and triplet losses to jointly supervise the training of the proposed deep model.

In recent a year, there are some literatures [9] [10-16] adopt part-based strategy to learn discriminative deep features for PReID. Cheng et al. [9] use a global convolution layer to get the global convolution features, and then they divide the features into four equal individual branches to obtain the part-based deep features. Finally, they concatenate the global and part-based feature vectors to produce the final deep features. In [10], the input images are first resized to 128×64 pixels. Then, the images are split into three overlapping parts and each of them with 64×64 pixels. Next, they use the three overlapping parts to input the three individual branches and use a fully connection layer to conclude the deep features of three branches. At last, the output deep feature vector is calculated by another fully connected layer. To handle the misalignment issue in person PReID, Zhao et al. [17] introduce a CNN-based attention model which utilizes the similarity information of a paired person images to learn the part body for matching. In [18], a Harmonious Attention Convolutional Neural Network is proposed to simultaneously learn feature representations and PReID selection in an end-to-end way. Specifically, they combine the hard attention and soft attention to learn the region-level and pixel-level parts of the person image, respectively. However, those part-based methods exist several limitations as follows: a). Introducing part branches to deep architecture increases the complexity of it, thus reducing the test efficiency. b). Ignoring the correlation between different part features. c). Ignoring the parts body's information from the horizontal orientation.

There also exist work try to exploit the relationship between channels. Hu et al. [19] introduce a mechanism which can integrate the network to perform feature recalibration. In this way, the informative features can be emphasized and the less useful one can be suppressed. In this paper, different from [19], we adopt GRU to learn the interdependencies knowledge between channels directly.

## III.. PROPOSED METHOD

Figure 2 shows the overall architecture of the proposed model. Our purpose is to learn omni-directional correlation information of pedestrian body. As can be seen, the backbone network adopted in this study is pre-trained ResNet-50 model, we reset the kernel size and stride to 1 at the layer4[0] of ResNet-50, for the higher spatial resolution before global pooling may contains more detail information of pedestrian features. Our network consists of five branches. It has been proved that the head, upper body and lower body as body part features learning can promote the discriminative ability of the learned pedestrian descriptors. Similar to other part-based feature learning model, we use branch-1 to learn the local part features from vertical orientation (from head to foot). As mentioned above, almost all of these part-based methods ignore to learn the part features of body from horizontal orientation. The horizontal orientation features may useful when facing the left or right side of bodies is occluded. Therefore, we utilize branch-2 to learn the part-based features from horizontal orientation. In addition to part-based features learning, we also introduce branch-3 to exploit the interdependencies knowledge between channels. The introduction of this branch can improve the robust ability of the learned deep features. Similar to our previous work [], the branch-4 and branch-5 are used to learn global features and a corresponding similarity measurement for PReID, respectively.

### A. Part-based features learning

The usual part-based approaches divide the pedestrian body into several horizontal stripes. Then each stripe is used to input an independent subnetwork to learn part feature from vertical orientation. Yet, these part-based methods ignore the spatial correlation between different stripes. Moreover, the part features learning from horizontal orientation is usually overlooked. To overcome these limitations above, in this paper, we use recurrent neural network with GRU cells to learn the spatial information between different local features both from vertical and horizontal orientations. As shown in Figure 2, we use the backbone network to extract deep features with dimension of $16\times8\times2048$. Then the deep features are implemented average pooling operation with $1\times8$, $1\times16$ kernel sizes in branch-1 and branch-2, respectively. As shown in Figure 3, each pooled deep features represents a corresponding receptive field from vertical or horizontal orientations. We use the pooled deep features as sequence features to feed into three layers bidirectional GRU, thus the two GRU branches can learn the part features with spatial information from vertical and horizontal orientations, respectively. We concatenate the outputted GRU features in branch-1 and branch-2 as the final part features, respectively. Finally, each of the two concatenated GRU features are used to input the two independent OIM layers to perform multi-classification task.

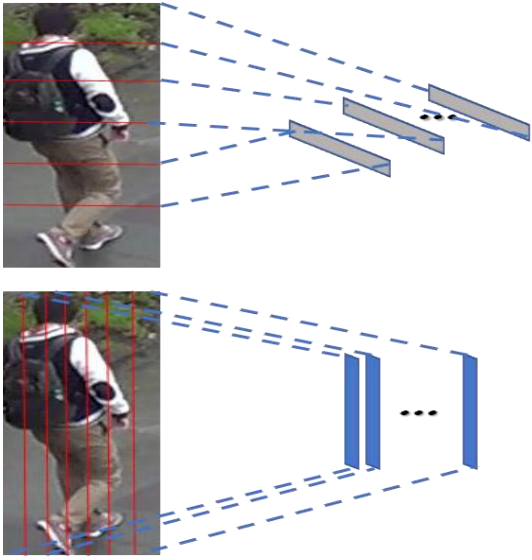

**Figure 3.** Each pooled deep feature vectors in branch-1 and branch-2 represents a corresponding receptive field of the pedestrian body from vertical and horizontal orientations, respectively.

### B. Relationship between Channels exploiting

The proposed model makes the effort to exploit the interdependencies between different channels as well. As shown in Figure 4, we convert each of feature maps with the size of 16×8 to a 128-dimension vector, thus each vector represents a channel. Then the 2048 vectors (channels) are used as the sequence features for feeding into three layers bidirectional GRU to learn the relationship between them. Finally, the outputs of GRU are also concatenated as a final descriptor to input the OIM layer. Through learning the interdependencies knowledge between these channels, we can improve the discriminatively ability of the learned representations.

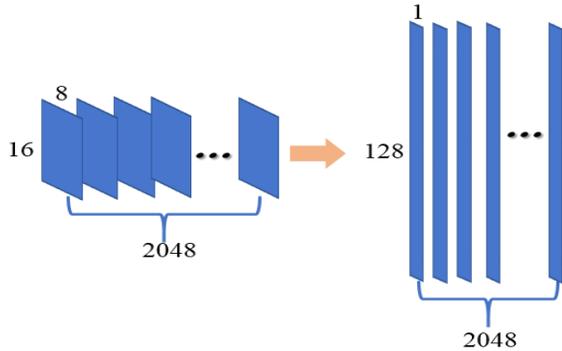

**Figure 4.** Each of 16×8 feature maps is converted to a 128-dimension vector.

### C. Loss functions

*Identification Loss.* In this work, different from most of existing multi-class recognition methods for PReID, we employ the Online Instance Matching (OIM) loss to train the identification sub-network. OIM loss is nonparametric, thus, the gradients can be performed on deep features without any classifier matrix transforming operations. The loss keeps a circular queue and a lookup table (LUT) to store the unlabeled and labeled features, respectively. The probability of pedestrian $M$ belonging to the identity $j$ can be formulated as:

$$p_j = \frac{\exp(v_j^T M / \partial)}{\sum_{i=1}^{L}\exp(v_i^T M / \partial) + \sum_{k=1}^{Q}\exp(\mu_k^T M / \partial)} \quad (1)$$

where $\mu_k^T$ and $v_j^T$ are the transpositions of the k-th identity of the circular queue and the j-th column of the LUT, respectively. The higher the temperature $\partial$ is, the softer the probability distribution. $Q$ and $L$ represent the queue size and the number of columns for the LUT, respectively.

The probability of being re-identified to the *j*-th *ID* in circular queue can be written as:

$$R_j = \frac{\exp(\mu_j^T M / \partial)}{\sum_{i=1}^{L}\exp(v_i^T M / \partial) + \sum_{k=1}^{Q}\exp(\mu_k^T M / \partial)} \quad (2)$$

The purpose of OIM is to maximize the log-likelihood:

$$L_{oim} = E_M[\log p_n] \quad (3)$$

*Triplet Loss.* For branch-5, we adopt the batch-hard triplet loss proposed by [20]. We randomly sample P identities (classes), and then choose $K$ images for each identity. Among these *PK* images, only the hardest positive and negative samples are selected for each anchor to form the triplet units. Given a triplet unit $(x_a, x_p, x_n)$, the loss function can be formulated as:

$$L_{trp} = \frac{1}{PK}\sum_{i=1}^{PK}[thre + \max_{+} d(F_w(x_a), F_w(x_p)) \quad (4)$$
$$- \min_{-} d(F_w(x_a), F_w(x_n))]_+$$

in which $[x]_+ = max(x,0)$, *thre* is a margin, $F_w(X)$ is the generated deep feature for image *X*, *W* represents the parameters of the backbone network. $d(x, y)$ is the distance between *x* and *y*. We adopt the Euclidean distance as metric function.

*Loss Function Combination.* The whole architecture is supervised by one triplet loss and four OIM losses. The final loss function is redefined as:

$$L = \lambda_1 L_{oim\_1} + \lambda_2 L_{oim\_2} + \lambda_3 L_{oim\_3} \\ + \lambda_4 L_{oim\_4} + \lambda_5 L_{trp} \quad (5)$$

where $\lambda_i$ $(i=1,2,3,4,5)$ represents trade-off parameter for different branch, $L_{oim\_i}$ $(i=1,2,3,4)$ denotes the OIM loss for different identification subnetwork.

## IV. EXPERIMENTS

We evaluate the proposed deep model on three widely used PReID benchmark datasets (i.e. CUHK03, Market-1501, and DukeMTMC_reID). Meanwhile, we present the results of some baseline configurations.

### A. Datasets and protocols

We use three well-known PReID datasets to evaluate the proposed model. The brief introductions of three datasets are presented as follows:

*CUHK03 dataset:* The dataset is one of the largest person ReID datasets which contains 13164 images of 1360 identities. All identities are taken from six camera views, and each pedestrian is captured by two cameras. This data set provides two setting. One automatically annotated by a detector and the other manually annotated by human. Among the two setting, the former is closer to practical scenarios. We both evaluate the proposed model on the two settings.

*Market-1501 dataset:* Market-1501 is a relatively large public available PReID dataset which is gathered from the front of a supermarket with a complex environment. It contains 32,668 images of 1501 IDs. Each of pedestrian taken from at least two cameras and at most six cameras.

*DukeMTMC-reID dataset*: The dataset created for image-based person ReID is a subset of DukeMTMC dataset. It consists of 36411 pedestrian images that belong to 1812 identities taken from eight high-resolution surveillance equipment. Among these 1812 pedestrians, 1404 of them captured by more than two camera views and the rest of them are regarded as distractor identifications.

*Evaluation metrics.* We use the Cumulative Match Characteristic (CMC) and mean Average Precision (mAP) as the evaluation protocols for each dataset. Both the evaluations on the three datasets are performed under single query setting.

### B. Implementation details

The model is performed on the pytorch framework. We discard the fully connected layer of the pre-trained ResNet-50 model and reset the kernel size and stride to 1 at the layer4 [0]. For branch-1 and branch-2, we set the hidden units of the GRUs to 256, and the hidden unites of GRU for branch-3 are set to 128. The training images are resized to 256×128. We use random erasing and random horizontal flipping as data augmentation for training. We set the *P* and *K* to 16 and 8, respectively. The training epochs are set to 150. We employ the Adaptive Moment Estimation (Adam) to optimize the parameters of the model. The weight decay and initial learning for Adam are set to $5\times10^{-4}$ and $2\times10^{-4}$, respectively. The learning rate is set according to the following update rules:

$$lr = \begin{cases} 2\times10^{-4} & \text{if epoch} \leq 100 \\ 2\times10^{-4} \times (0.001^{((\text{epoch}-100)/50)}) \end{cases} \quad (6)$$

We set the all $\lambda_i$ $(i=1,2,3,4,5)$ to 1. The margin *thre* is set to 0.5. During the test phase, the images are also resized to 256×128. We use the feature $f_{OIM}$ as descriptor to retrieval for Market-1501 and DukeMTMC-reID datasets. As to the CUHK03 dataset, the feature $f_{Trip}$ is used for retrieving.

### C. Comparison with state-of-the-art methods

We compare the proposed architecture with other state-of-the-art methods, including DNS, Gated Siamese, HydraPlus-Net, CNN+DCGAN, SVD-Net, Deep Transfer, TriNet, PCB, Deep-Person, PCB+RPP, Part-Aligned, JLML, PDC, MSCAN, GOG, SSM, LSRO, BoW + KISSME, OIM, PAN, ACRN, DPFL, SPReID. Among these comparison methods, SPReID achieves the highest performance for PReID in the CVPR 2018. The comparison details are presented as follows:

**Table 1.** Comparison results on the Market-1501 dataset. '-' means no available reported results.

| | Single Query | | | |
|---|---|---|---|---|
| **Method** | **mAP** | **Rank-1** | **Rank-5** | **Rank-10** |
| DNS [21] | 35.6 | 61.0 | - | - |
| Gated Siamese [22] | 39.5 | 65.8 | - | - |
| HydraPlus-Net [23] | - | 76.9 | 90.9 | - |
| CNN+DCGAN [24] | 56.2 | 78.0 | - | - |
| CNN Embedding [25] | 59.8 | 79.5 | 92. | 95.2 |
| SVD-Net [26] | 62.1 | 82.3 | 91.3 | 94.5 |
| Deep Transfer [27] | 65.5 | 83.7 | - | - |
| TriNet [20]+RK | 81.0 | 86.6 | 93.3 | - |
| CamStyle [28]+RK | 71.5 | 89.4 | - | - |
| HA-CNN [29] | 75.7 | 91.2 | - | - |
| PCB [30] | 77.4 | 92.3 | 97.2 | 98.2 |
| Deep-Person [2] | 79.6 | 92.3 | - | - |
| PCB+RPP[30] | 81.6 | 93.8 | 97.5 | 98.5 |
| SPReID [31] | 83.3 | 93.6 | 97.5 | 98.4 |
| Our | **85.4** | **94.7** | **98.6** | **99.0** |

*Evaluation on Market-1501 dataset.* For this dataset, We follow the dataset setting and use 12936 images of 751 identities for training. The rest 19732 images of 750 identities are used for test. The comparison results of our model aginst fourteen methods on Market-1501 are presented in Table 1. We can obesrve that the proposed model outperforms the all competing approaches, which demonstrates the effectiveness of the proposed method. Specially, we achieve mAP= 85.4%, Rank-1 accuracy = 94.7%, Rank-5 accuracy= 98.6% and Rank-10 accuracy = 99.0% under signle query mode.

**Table 2.** Comparison results on the CUHK03_labeled dataset. '-' means no available reported results.

| **Method** | **Rank-1** | **Rank-5** | **Rank-10** |
|---|---|---|---|

| Method | Rank-1 | Rank-5 | Rank-10 |
|---|---|---|---|
| Siamese LSTM [1] | 57.30 | 80.10 | 88.30 |
| GOG [32] | 67.30 | 91.00 | 96.00 |
| DGD [30] | 80.50 | 94.90 | 97.10 |
| Quadruplet [33] | 74.47 | 96.62 | 98.95 |
| CNN Embedding [25] | 83.40 | 97.10 | 98.70 |
| Deep Transfer [27] | 84.10 | - | - |
| Part-Aligned [34] | 85.4 | 97.6 | 99.4 |
| MLS Deep [35] | 87.5 | 97.8 | 99.4 |
| Deep-Person [2] | 91.5 | 99.0 | 99.5 |
| Deep CRF [36] | 90.2 | 98.5 | - |
| HydraPlus [23] | 91.8 | 98.4 | 99.1 |
| SPReID [31] | 94.2 | 99.0 | 99.5 |
| Our | **95.0** | **99.3** | **99.5** |

Table 3. Comparison results on the CUHK03_detected dataset. '-' means no available reported results.

| Method | Rank-1 | Rank-5 | Rank-10 |
|---|---|---|---|
| DNS [21] | 54.7 | 84.7 | 94.8 |
| GOG [32] | 65.5 | 88.4 | 93.7 |
| Gated Siamese [22] | 68.1 | 88.1 | 94.6 |
| SVD-Net [26] | 81.8 | 95.2 | 97.2 |
| Part-Aligned [34] | 81.6 | 97.3 | 98.4 |
| JLML [37] | 80.6 | 96.9 | 98.7 |
| MLS Deep [35] | 86.4 | 97.5 | 99.1 |
| Deep CRF [36] | 88.8 | 97.2 | - |
| Deep-Person [2] | 89.4 | 98.2 | 99.1 |
| Our | **92.0** | **98.9** | **99.5** |

*Evaluation on CUHK03 dataset.* For CUHK03 dataset, we select 1160 identities for training and 100 identities for validation as well as 100 identities for test. Table 2 and Table 3 show the comparison results on two setting CUHK03 datasets. From Table 2 and Table 3, we can see that the performances of the proposed model is superior to all other compared methods on both labeled and detected CUHK03 datasets. Our proposed model yields Rank-1 accuracy = 95.0%, Rank-5 accuracy= 99.3% on CUHK03_labeled dataset and Rank-1 accuracy = 92.0%, Rank-5 accuracy= 98.9% on CUHK03_detected dataset

Table 4. Comparison results on the DukeMTMC-reID dataset. '-' means no available reported results.

| Method | mAP | Rank-1 | Rank-5 | Rank-10 |
|---|---|---|---|---|
| LOMO+XQDA [38] | 17.0 | 30.7 | - | - |
| CNN+DCGAN [39] | 47.1 | 67.6 | - | - |
| PAN [40] | 51.5 | 71.5 | - | - |
| OIM [3] | 47.4 | 68.1 | - | - |
| CNN embedding [25] | 49.3 | 68.9 | - | - |
| SVD-Net [26] | 56.8 | 76.7 | 86.4 | 89.9 |
| TriNet [20] | 53.5 | 72.4 | - | - |
| ACRN [41] | 51.9 | 72.5 | 84.7 | - |
| Deep CRF [36] | 69.5 | 84.9 | - | - |
| Deep-Person [2] | 64.8 | 80.9 | - | - |
| SPReID [31] | 73.3 | 85.9 | 92.9 | 94.5 |
| Our | **74.6** | **86.7** | **93.4** | **95.7** |

*Evaluation on DukeMTMC-reID dataset.* The 16,522 images of 702 persons of this dataset are used for training, and the remaining 702 persons are divided into query images and gallery images. As shown in Table 4, the proposed method achieves mAP= 74.6%, Rank-1 accuracy = 86.7%, Rank-5 accuracy= 93.4% and Rank-10 = 95.7%, which outperforms the compared state-of-the-art methods and further verifies the effectiveness of the proposed method.

### D. Ablation analysis and discussions

To evaluate the effectiveness of each component in the proposed model, we design several baseline models with different configurations to make comparisons between them. The details of comparison experiments are presented as below:

*Effect of GRU-based part features learning.* In this setting, we first discard the GRU-based part features learning branches, *i.e.* Branch-1 and Branch-2, and train the network with the rest branches. We name this configuration model as Our-G. Then we add Branch-1 to the Our-G and name it as Our-G-B1. Finally, the Branch-2 is added to Our-G-B1. As depicted in Table 5, among the three configurations, Our-G gets the worst results. When we add Branch-1 to it, the performance of Our-G is promoted obviously. The model Our which integrates the two branches achieves the best performance among the three configurations, which demonstrates the effectiveness of the GRU-based part features learning branches.

Table 5. The effectiveness of GRU-based part features learning

| Model | mAP | Rank-1 | Rank-5 |
|---|---|---|---|
| Our-G | 82.1 | 92.8 | 98.3 |
| Our-G-B1 | 83.5 | 93.6 | 98.2 |
| Our | **85.4** | **94.7** | **98.6** |

*Effect of relationship between channels exploiting.* In this setting, the Branch-3 of the proposed model is discard. we name this discarded model as Our/$_{channels}$. From Table 6, we can observe that the performance of Our/$_{channels}$ is enhanced when introducing the Branch-3, which proves that the interdependencies knowledge between different channels learning can improve the discriminatively ability of the learned descriptor.

Table 6. The effectiveness of the relationship between channels exploiting

| Model | mAP | Rank-1 | Rank-5 |
|---|---|---|---|
| Our/$_{channels}$ | 84.1 | 93.8 | 98.4 |
| Our | **85.4** | **94.7** | **98.6** |

*Person descriptor choice.* As mentioned above, we use the feature $f_{Trip}$ as the person descriptor to re-identify for CUHK03_labeled dataset. As for the Market-1501 and DukeMTMC-reID datasets, we choose the feature $f_{OIM}$ to retrieve. In order to validate the validity of the selection strategy, we also use the feature $f_{Trip}$ to retrieve for Market-1501 and DukeMTMC-reID datasets and use the feature $f_{OIM}$ to retrieve for CUHK03_labeled dataset. As shown in Table 7, when use the feature $f_{OIM}$ as descriptor for CUHK03, the accuracies of mAP and Rank-1 obviously drop. We speculate this is due that the CUHK03 dataset contains a large number of identities and each identity only has limited number image samples (average 9.6). Using such scale datasets to train the multi-classification branch-3 may introduces over-fitting issue, so we use the feature $f_{Trip}$ as the final representation for CUHK03 dataset. Different from [2], compared with feature $f_{Trip}$, $f_{OIM}$ achieves slight higher results on both Market-1501 and DukeMTMC-reID datasets. This indicates that OIM loss may achieve better performance in relatively larger PReID dataset for our deep architecture, which demonstrates the effectiveness of our choice strategy.

**Table 7.** The effectiveness of the feature selection strategy

| dataset | Feature | mAP | Rank-1 |
|---|---|---|---|
| CUHK03_labeled | $f_{OIM}$ | 88.4 | 91.2 |
| | $f_{Trip}$ | 92.2 | 95.0 |
| Market-1501 | $f_{OIM}$ | 85.4 | 94.7 |
| | $f_{Trip}$ | 84.1 | 93.6 |
| DukeMTMC-reID | $f_{OIM}$ | 74.6 | 86.7 |
| | $f_{Trip}$ | 74.0 | 85.4 |

## VI. CONCLUSIONS

In this study, we propose an omni-directional feature learning deep model for PReID. The model can learn part representation with spatial information from vertical and horizontal orientations. To further improve the discriminatively ability of the learned descriptors, we use GRU to exploit the relationship between channels. Moreover, the triplet loss and OIM loss are employed to learn a similarity measurement and global features at the same time. Extensive experiments on CUHK03, Market-1501 and DukeMTMC-reID datasets show the proposed deep architecture achieves state-of-the-art results.